\documentclass[sigconf]{acmart}

\newcommand{\ie}{\textit{i}.\textit{e}.}
\newcommand{\eg}{\textit{e}.\textit{g}.}
\newcommand{\cf}{\textit{cf.}}
\newcommand{\etal}{\textit{et}.\textit{al}.}

\usepackage{amsmath}
\usepackage{amsthm}
\usepackage{mathtools}
\usepackage{tabularx}
\usepackage{multirow}
\usepackage{xcolor}
\usepackage{colortbl}
\usepackage{tablefootnote}

\usepackage{graphicx}
\usepackage{subcaption}
\usepackage{bm}
\usepackage{pifont}  
\usepackage{hhline}
\usepackage{enumitem}
\usepackage{etoolbox}
\usepackage{listings}
\makeatletter  
\newcommand{\thickhline}{%
    \noalign {\ifnum 0=`}\fi \hrule height 1pt
    \futurelet \reserved@a \@xhline
}

\definecolor{mygray}{gray}{.9}
\definecolor{ggray}{RGB}{127,127,127}
\definecolor{reda}{RGB}{192,0,0}
\definecolor{redb}{RGB}{217,148,143}
\definecolor{myyellow}{RGB}{190,144,0}
\definecolor{mygreen}{RGB}{0,176,80}
\definecolor{myred}{RGB}{248,66,0}
\definecolor{myblue}{RGB}{30,90,100}
\definecolor{mygray1}{RGB}{245,245,245}

\definecolor{agent}{RGB}{218, 165, 32}
\definecolor{goods}{RGB}{220, 20, 60}
\definecolor{payment}{RGB}{34,139,34}
\definecolor{seller}{RGB}{70,130,180}
\definecolor{place}{RGB}{138,43,226}

\theoremstyle{definition}

\AtBeginDocument{%
  }

\copyrightyear{2025}
\acmYear{2025}
\setcopyright{acmlicensed}\acmConference[MM '25]{Proceedings of the 33rd
ACM International Conference on Multimedia}{October 27--31, 2025}{Dublin,
Ireland}
\acmBooktitle{Proceedings of the 33rd ACM International Conference on
Multimedia (MM '25), October 27--31, 2025, Dublin, Ireland}
\acmDOI{10.1145/3746027.3755134}
\acmISBN{979-8-4007-2035-2/2025/10}
\begin{document}

\title[Zero-shot Compositional Action Recognition with Neural Logic Constraints]{Zero-shot Compositional Action Recognition \\ with Neural Logic Constraints}

\author{Gefan Ye}
\orcid{0009-0009-3030-405X}
\authornote{Both authors contributed equally to this research.}
\affiliation{%
\department{College of Computer Science and Technology}
  \institution{Zhejiang University}
  \city{Hangzhou}
  \country{China}}
\email{gefanyeh@zju.edu.cn}

\author{Lin Li}
\orcid{0000-0002-5678-4487}
\authornotemark[1]
\authornote{Lin Li is the corresponding author.}
\affiliation{%
  \institution{
  AI Chip Center for Emerging Smart Systems
  }
  \city{Hong Kong}
  \country{China}}
\affiliation{%
  \institution{The Hong Kong University of Science and Technology}
  \city{Hong Kong}
  \country{China}}
\email{lllidy@ust.hk}

\author{Kexin Li}
\orcid{0009-0000-0726-0196}
\authornotemark[1]
\affiliation{%
  \institution{Zhejiang Tobacco Monopoly Administration}
  \city{Hangzhou}
  \country{China}}
\email{likexin@zj.com.yc}

\author{Jun Xiao}
\orcid{0000-0002-6142-9914}
\affiliation{%
\department{College of Computer Science and Technology}
\institution{Zhejiang University}
\city{Hangzhou}
\country{China}}
\email{junx@cs.zju.edu.cn}

\author{Long Chen}
\orcid{0000-0001-6148-9709}
\affiliation{%
  \institution{The Hong Kong University of Science and Technology}
  \city{Hong Kong}
  \country{China}}
\email{longchen@ust.hk}

\renewcommand{\shortauthors}{Gefan Ye, Lin Li, Kexin Li, Jun Xiao, \& Long Chen}

\begin{abstract}
Zero-shot compositional action recognition (ZS-CAR) aims to identify unseen verb-object compositions in the videos by exploiting the learned knowledge of verb and object primitives during training. Despite compositional learning's progress in ZS-CAR, two critical challenges persist: 1) \textit{Missing compositional structure constraint}, leading to spurious correlations between primitives; 2) \textit{Neglecting semantic hierarchy constraint}, leading to semantic ambiguity and impairing the training process. In this paper, we argue that human-like \textit{symbolic reasoning }offers a principled solution to these challenges by explicitly modeling compositional and hierarchical structured abstraction. To this end, we propose a logic-driven ZS-CAR framework \textbf{LogicCAR} that integrates dual symbolic constraints: \textbf{Explicit Compositional Logic} and \textbf{Hierarchical Primitive Logic}. Specifically, the former models the restrictions within the compositions, enhancing the compositional reasoning ability of our model. The latter investigates the semantical dependencies among different primitives, empowering the models with fine-to-coarse reasoning capacity. By formalizing these constraints in first-order logic and embedding them into neural network architectures, LogicCAR systematically bridges the gap between symbolic abstraction and existing models.
Extensive experiments on the Sth-com dataset demonstrate that our LogicCAR outperforms existing baseline methods, proving the effectiveness of our logic-driven constraints.
\end{abstract}



\maketitle

\section{Introduction}

\begin{figure}[!t]
   \begin{center}
       \includegraphics[width=\linewidth]{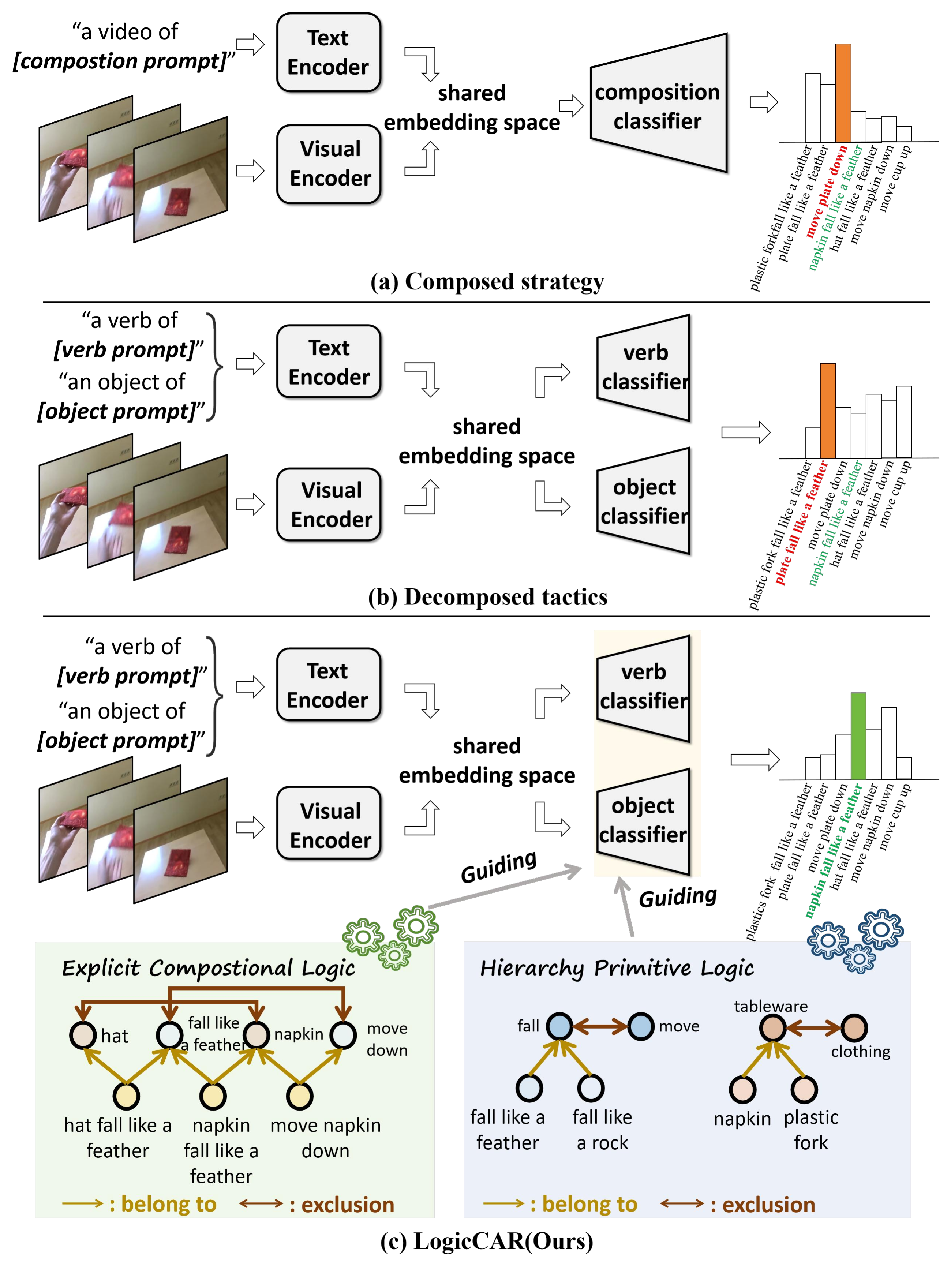}
   \end{center}
   \vspace{-2.0em}
   \captionsetup{font=small}
   \caption{Illustrations of previous compositional learning methods and LogicCAR (ours). Take a video sample of ``\textit{napkin fall like a feather}'' as an example, ours (\ie, (c)) integrates symbolic reasoning into decomposed tactics (\ie, (b)), leading to an obvious accuracy improvements compared with other methods (\ie, (a), (b)). }
   \vspace{-10pt}
   \label{fig:motivation}
\end{figure}

We humans exhibit an inherent talent for comprehending new concepts. This can be attributed to our extraordinary ability to generalize from what we have learned before. For instance, when first faced with the concept ``\textit{fold napkin}'', we are able to imagine how the action looks like by combining the verb ``\textit{fold}'' with the object ``\textit{napkin}'' from familiar concepts ``\textit{fold sock}'' and ``\textit{tear napkin into two pieces}''. Inspired by
this native cognitive ability of humans, Zero-Shot Compositional Action Recognition (ZS-CAR)~\cite{li2024c2c} emerges to tackle \textit{unseen} verb-object pairs recognition based on \textit{seen} verb primitives and object primitives in the video understanding field. Compared with image-level compositional zero-shot learning~\cite{nan2019recognizing, zheng2024caila, jayasekara2024unified}, ZS-CAR poses a finer-grained challenge regarding exploring information about spatial and temporal dynamics in videos.

Naturally, it is intuitive to incorporate compositional learning approaches~\cite{kato2018compositional,lu2023decomposed,huang2024troika} to deal with the challenging ZS-CAR task. Existing compositional learning methods for classification fall broadly into two categories: 1) Composed strategy~\cite{kato2018compositional, jayasekara2024unified, nagarajan2018attributes, nan2019recognizing}: It specializes in the whole composition classification through shared feature space projection, where integrated visual attributes are jointly embedded (\cf, Figure~\ref{fig:motivation}(a)); 2) Decomposed tactics~\cite{hou2020visual, zheng2024caila, li2024compositional}: It decouples visual representations of elementary primitives through a dual-branch classifier architecture that independently predicts two different primitives, as illustrated in Figure~\ref{fig:motivation}(b).

Despite their effectiveness, these methods still face two main obstacles: 1) \textbf{Missing compositional structure constraint}: Current compositional learning methods fail to explicitly model compositional relationships by merely concatenating primitive embeddings through naive additive fusion (\eg, late fusion of individual verb and object classifiers). It can lead to the spurious correlation that the model may learn incorrect verb-object associations 
(\eg, predicting ``\textit{napkin fall like a rock}'' as valid composition if ``\textit{fall like a rock}'' and ``\textit{napkin}'' frequently co-occur in the training data). 
2) \textbf{Neglecting semantic hierarchy constraint}: These approaches equally treat all primitive labels and ignore the semantic hierarchy. It results in semantic confusion and makes it hard to establish proper decision boundaries. Particularly in the spatiotemporal domain, neglecting semantic hierarchy constraints exacerbates challenges due to complex inter-frame dependencies. For example, the motion trajectories for the verb ``\textit{fall like a feather}'' and the verb ``\textit{fall like a rock}'' apparently share certain commonalities. Therefore, ignoring the semantic hierarchy between the two verbs hinders models from developing a deeper insight into the action ``fall''.

Nevertheless, as humans, we possess not only the ability for compositional reasoning but also the capacity for fine-to-coarse reasoning within hierarchical structures.
To elaborate, we can effortlessly ascertain the rationality of the compositional structure of verb-object pairs. With reference to our common sense, it is a trivial matter to figure out whether one verb category and one object category manifest a logical inconsistency between them. Take the case of the verb ``fall like a rock'' as an example. It is universally acknowledged for us humans that a valid composition for ``\textit{fall like a rock}'' must come with a heavy object. Objects such as ``\textit{napkin}'' and ``\textit{paper}'' are in conflict with the outlined conditions, resulting in ``\textit{napkin fall like a rock}'' and ``\textit{paper fall like a rock}'' being deemed unreasonable compositions. 
Furthermore, human linguists can use hierarchical dependencies to categorize verb categories and object categories into different groups (\ie, coarse verb categories and coarse object categories), respectively. Words in the same group generally have semantic similarities and even make substitutions in certain cases. With the knowledge of such logical coherence among coarse categories, it is a piece of cake for us to better understand a new concept and judge its reasonableness. For instance, with the preliminary that ``\textit{hat}'' and ``\textit{shoe}'' are in the coarse object category of ``\textit{clothing}'', one can readily notice that ``\textit{wear hat}'' and ``\textit{wear shoe}'' are both plausible compositions. 

In this paper, inspired by such ``logical-thinking'' capacity, we propose \textbf{LogicCAR}, a logical reasoning framework for ZS-CAR by leveraging the symbolic knowledge expressed in the form of first-order logic, as displayed in Figure~\ref{fig:motivation}(c). To elaborate, LogicCAR mimics ``human-like'' symbolic reasoning by establishing compositional and hierarchical structured abstractions among verb and object primitives. This process integrates two core principles:

\textbf{Explicit Compositional Logic}. It models the compositional connections in an explicit way, which contain: 1) \textit{composed} relationships between compositions and their verb categories (\eg, composition category ``\textit{napkin fall like a feather}'' and verb category ``\textit{fall like a feather}''), and \textit{composed} relationships between compositions and their object categories (\eg, composition category ``\textit{napkin fall like a feather}'' and object category ``\textit{napkin}''); 2) \textit{exclusive} relationships among different verb categories (\eg, verb category ``\textit{fall like a feather}'' and verb category ``\textit{move down}''), and \textit{exclusive} relationships among different object categories (\eg, object category ``\textit{napkin}'' and object category ``\textit{hat}'').

\textbf{Hierarchical Primitive Logic}. It emphasizes the dependencies among verb categories and object categories,
including: 1) \textit{composed} relationships between verb categories and
their coarse verb categories (\eg, verb category ``\textit{fall like a feather}'' and coarse verb category ``\textit{fall}''), and \textit{composed} relationships between
object categories and their coarse object categories (\eg, object category ``\textit{napkin}'' and coarse object category ``\textit{tableware}''); 2) \textit{exclusive}
relationships among different coarse verb categories (\eg, coarse verb category ``\textit{fall}'' and coarse verb category ``\textit{move}''), and \textit{exclusive} relationships among different coarse object categories (\eg, coarse object category ``\textit{tableware}'' and coarse object category ``\textit{clothing}'').

To integrate these logic constraints with existing neural networks, we first formalize them using \textit{first-order logic}~\cite{barwise1977introduction, fischer2019dl2} to encode semantic dependencies as predicate-based axioms. These axioms are then relaxed via \textit{fuzzy logic}~\cite{kosko1993fuzzy, van2022analyzing} into differentiable loss terms, enabling seamless integration into gradient-based optimization during training. 
By leveraging these formalized logical rules, LogicCAR simultaneously enhances generalization capability in unseen scenarios through its compositional and hierarchical structures and maintains high interpretability by explicitly encoding symbolic knowledge as transparent constraints.

To verify the effectiveness of our work, we conduct a series of experiments and ablation studies on Sth-com dataset. It proves that our logic-driven constraints contribute to the improvement of the comprehensive performance on ZS-CAR (ours, \textit{AUC} \textbf{27.0\%} and \textit{HM} \textbf{45.2\%}), compared with the state-of-the-art methods (\eg, C2C, \textit{AUC} 25.9\% and \textit{HM} 44.8\%). The code will be released soon.

To summarize, the main contributions of our work are as follows:
\begin{itemize}[leftmargin=*]
    \item We analyze the prevailing limitations in directly applying existing compositional learning methods in ZS-CAR task, \ie, missing both compositional and semantic hierarchy constraints.
    \item We propose LogicCAR, a logic reasoning framework with dual logic-driven constraints to enhance the understanding of compositional actions in videos: explicit compositional logic and hierarchical primitive logic. To the best of our knowledge, it is the first work to integrate neural-symbolic reasoning in ZS-CAR.
    \item Extensive results on the Sth-com dataset demonstrate that LogicCAR outperforms existing methods, showcasing its effectiveness in recognizing \textit{unseen} verb-object compositions in ZS-CAR task.
\end{itemize}

\section{Related Work}

\noindent\textbf{Compositional Action Recognition (CAR).} With rapid advances in deep learning, models perform well in the field of video action recognition~\cite{yao2019review, pareek2021survey, ahmad2021graph, koohzadi2017survey, shao2022deep}. To facilitate models' generalization ability, Materzynska et al.~\cite{materzynska2020something} propose a new task: compositional action recognition. It requires recognizing the verb for verb-object pairs in the videos whereas the verb is performed with objects that do not exist together with this verb in the training phase. To evaluate models' performance on the task, they construct a new benchmark named Something-Else, based on Something-Something V2 dataset. Action Genome~\cite{ji2020action} decomposes compositional actions into spatio-temporal scene graphs and focuses on the dynamics of compositional actions. It also contributes to few-shot compositional action detection. LLTM~\cite{yan2022look} executes mutations on objects in the video to reduce the strong link between visual objects and action-level labels, and aligns different modality information in the contrastive embedding space to learn the commonsense relationships. CDN~\cite{sun2021counterfactual} makes use of factual inference and counterfactual inference to conduct a causal graph with no bias of visual appearance information, thus improving generalization ability. However, previous methods still struggle to truly tackle the challenge of generalization for compositions, leading to unsatisfactory results when it comes to recognizing both verbs and objects.

\noindent\textbf{Compositional Zero-Shot Learning (CZSL).} CZSL~\cite{mancini2021open, mancini2022learning, li2022siamese} concentrates on predicting unseen combinations of attributes and objects while individual primitives are seen during the training phase. Generally speaking, current CZSL frameworks can be divided into two groups: 1) \textbf{Decomposed CZSL}~\cite{zhang2022learning, hao2023learning, khan2023learning, li2020symmetry, saini2022disentangling, liu2023simple}: It disentangles embeddings of attributes and objects via using two classifiers to predict them respectively. 2) \textbf{Composed CZSL}~\cite{purushwalkam2019task, anwaar2022leveraging, wei2019adversarial}: It treats attribute-object pairs as a whole and aligns compositions with images in a shared embedding space. Over recent years, both decomposed and composed CZSL have witnessed remarkable breakthroughs with the aid of CLIP~\cite{huang2024troika, nayak2022learning, lu2023decomposed, wang2023hierarchical}. Pretrained vision-language models enable CZSL approaches to capture complex compositional semantics, thus enhancing generalization ability. Nonetheless, prior work fails to delve deeply into the internal architecture of compositions, neglecting the inherent connection between compositions and primitives. 

\noindent\textbf{Neuro-Symbolic Computing (NSC).} NSC refers to a technology stipulating that knowledge is expressed in symbolic representations and employs a neural network to compute learning and reasoning~\cite{garcez2019neural}. Although current NSC systems usually provide aid to AI models, the origins of NSC research can be traced back to 1943~\cite{mcculloch1943logical}, long before AI emerged as a pivotal research area. Early NSC systems focus on how to effectively combine symbolic learning and neural learning~\cite{shavlik1994combining, mcgarry1999hybrid, sun1994computational, tan1997cascade}. Regrettably, hampered by insufficient device resources and crude design of internal interfaces, NSC in that era failed to fully harness its potential. Currently, NSC systems combined with tactics like propositional logic~\cite{francca2014fast, tran2016deep}, first-order logic~\cite{pitangui2012learning, serafini2016learning, gori2023machine} and temporal logic~\cite{borges2011learning, de2011neural} offer foundation models ground-breaking solutions to further enhance the reasoning ability.  

\noindent\textbf{Large Language Models (LLMs).} LLMs have seen tremendous advancements recently, demonstrating strong capabilities in a wide range of tasks in the natural language processing (NLP) domain. According to the architectural design and main functionality of LLMs, there exist three types of LLMs: 1) Encoder-Only LLMs~\cite{devlin2019bert, liu2019roberta, lan2019albert}: This kind of LLMs emphasizes natural language understanding and has good behavior in the field of classification and extraction. With the bidirectional encoder as the core of model structure, encoder-only LLMs create dynamic contextual embeddings for every token, enhancing the comprehension of input content. 2) Encoder-Decoder LLMs~\cite{raffel2020exploring, lewis2019bart, song2019mass}: Compared with encoder-only LLMs, this type of LLMs adds a decoder in the framework. By leveraging contextual representations from the encoder, the decoder auto-regressively generates the target sequence, boosting LLMs' effectiveness in addressing sequence-to-sequence tasks, such as translation, generation, and so on. 3) Decoder-only LLMs~\cite{radford2018improving, radford2019language, achiam2023gpt, touvron2023llama, chowdhery2023palm, bi2024deepseek, liu2024deepseek, li2024collaborative}: Decoder-only LLMs operate exclusively with a Transformer decoder, forgoing any encoder components. Such an architecture speeds up the inference process and offers strong scalability, empowering decoder-only LLMs to excel in conversation tasks. Nowadays, with the support of techniques like multi-modal learning~\cite{zhang2024mm, wang2024exploring, gong2023multimodal, li2023catr, lil2024survey, Chen_2025_CVPR} and parameter-efficient fine-tuning~\cite{han2024parameter, gao2023llama, li2023prompt, li2024idpro}, LLMs have nearly attained a human-expert level across a wide range of tasks in the NLP domain. Consequently, it is no wonder that LLMs have established themselves as a significant assistant in NLP field.

More related works are left in the \textbf{Appendix}.
\section{Methodology}

\begin{figure*}[t]
   \vspace{-10pt}
   \begin{center}
      \includegraphics[width=\linewidth]{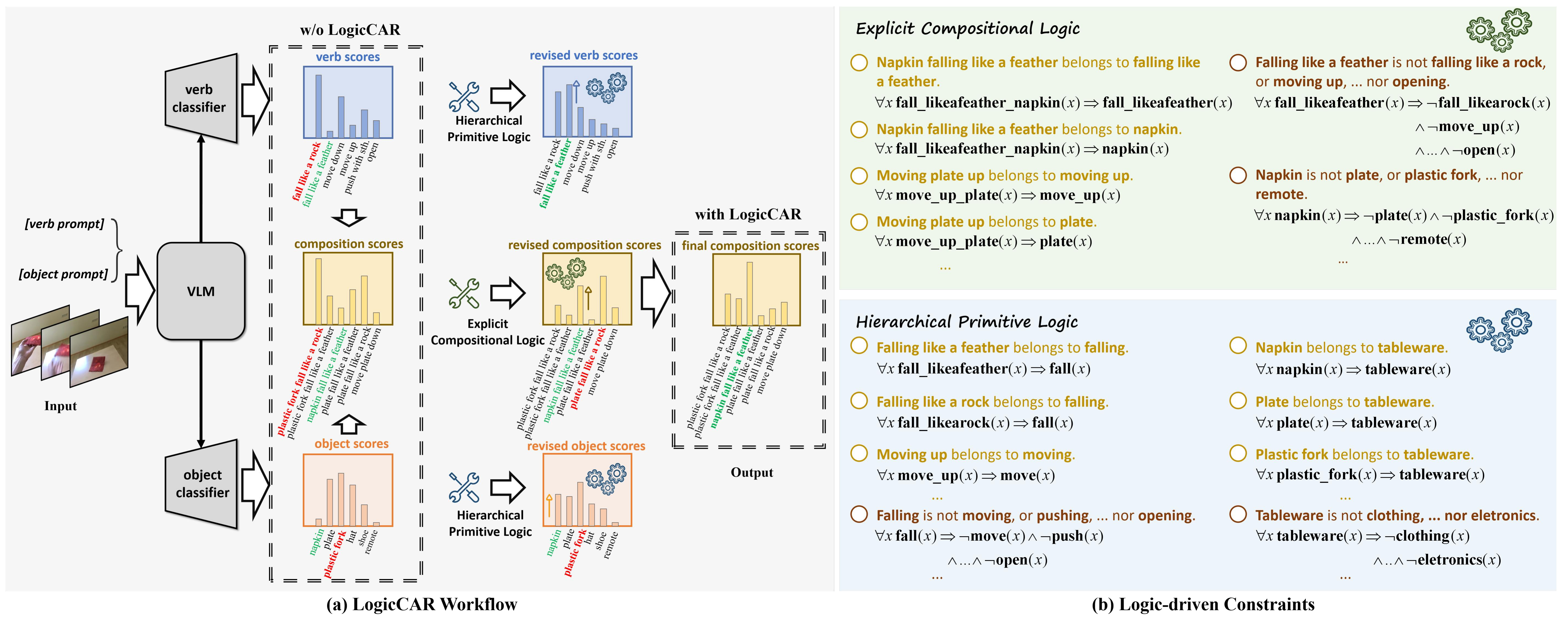}
   \end{center}
   \vspace{-18pt}
   \captionsetup{font=small}
   \caption{An overview of LogicCAR: a framework with two logic-driven constraints for ZS-CAR task. Explicit Compositional Logic investigates the restrictions within compositions, while Hierarchical Primitive Logic focuses on the semantic hierarchy of verb categories and object categories. The workflow is shown in (a), while some specific rules in logic-driven constraints are detailed in (b).}
   \vspace{-5pt}
   \label{fig:overview}
\end{figure*}

\textbf{Problem Formulation.} Given a video sample $V$, ZS-CAR aims to identify videos based on composition categories $c \in C$, where the composition category $c$ is defined as a verb-object pair: $c =(v, o)$. All verb categories  $v$ come from a verb set $V$ = $\{v_1, \ldots, v_{|V|}\}$, and similarly all object categories $o$ come from an object set $O$ = $\{o_1, \ldots, o_{|O|}\}$. Restricted by feasible compositions in the real world, the category set $C$ is only a subset of the Cartesian product of the verb set $V$ and object set $O$, \ie, $C \in V \times O$. \textbf{For zero-shot scenarios}, when in the test period, the model has to recognize compositions that never occur during training. That is to say, in the training phase, the model can only access a subset of the whole category set $C$, \ie, a set of seen compositions, $C(s)$. For notational convenience, we denote the set of unseen compositions as $C(u)$. 

In this section, we first revisit ZS-CAR baselines in Sec.~\ref{3.1}. Then, we explain the details about our two logic-driven constraints in LogicCAR, \ie, Explicit Compositional Logic and Hierarchical Primitive Logic, in Sec.~\ref{3.2}. Subsequent to this, we present a description of our training objective in Sec.~\ref{3.3}.

\subsection{Revisit ZS-CAR Baselines}
\label{3.1}
The general ZS-CAR framework consists of two types of encoders: visual encoders $En_V(\cdot)$ and text encoders $En_T(\cdot)$. With the help of these encoders, the framework enables the joint projection of textual and visual data into a shared embedding space and learns cross-modal associations. Therefore, the framework is capable of recognizing unseen verb-object compositions via what it has learned in the training.

\textbf{1) Visual Encoders $En_V(\cdot)$.} 
They extract visual features of the input video. Considering the spatial-temporal characteristics of video data, visual encoders~\cite{wang2021actionclip, wang2023seeing} are specifically designed to capture both the detailed spatial information within individual frames and the temporal dynamics across sequential frames. In this paper, we follow the~\cite{li2024c2c}, utilizing extra \textit{Dynamic Module} and \textit{Static Module} to disentangle the temporal dynamic features of verb primitives and the static visual information of object primitives, respectively. The former consists of a temporal pooling layer and two MLP layers, while the latter utilizes two temporal convolution layers and a temporal pooling layer. Specifically, we first extract general representations $\mathbf{F}_x \in \mathbb{R}^{T \times D}$ of the video frames, with $T$ as the frame number and $D$ as the channel number. Then \textit{Dynamic Module} and \textit{Static Module} leverage $\mathbf{F}_x$ to obtain dynamic visual features (\ie, verb-related visual features) and static visual features (\ie, object-related visual features), respectively.

\textbf{2) Text Encoders $En_T(\cdot)$.} They extract text features from verb-contained or object-contained prompts (\eg, ``a verb of x'' or ``an object of x'').
To adapt to specific tasks, Parameter-Efficient Fine-Tuning (PEFT)~\cite{han2024parameter} strategies are applied, such as inserting small trainable modules (Adapters) into Transformer layers, updating only a few parameters while keeping pre-trained weights frozen~\cite{zheng2024caila, huang2024troika}. Another approach, Prompt Tuning~\cite{lu2023decomposed, zhou2022conditional, nayak2022learning}, optimizes learnable prompts to guide the encoder to focus on task-relevant features like verb-object relationships.
These PEFT methods enable efficient handling of downstream tasks with minimal computational overhead. To elaborate, in this paper, we incorporate the techniques of adapters and prompt tuning to optimize our text encoders. Meanwhile, to discover the optimal prompt-tuning protocol, we design several experiments to compare the impacts of different prompt strategies on LogicCAR.

\textbf{3) Embedding-based Classification.} With text embeddings and visual features, models are then able to start to study their cross-modal relationships via contrastive learning~\cite{radford2021learning, wang2021actionclip, zhou2022conditional}. This learning approach leverages cosine similarities between textual and visual representations as the determining factor in predictions. During training, it optimizes the loss of cosine similarities to learn the correct mapping correlations between texts and videos. And in the test phase, the softmax function is conventionally the default activation to derive final predictions from cosine similarities. In this paper, we follow the above pipeline of contrastive learning to implement multi-label classification for ZS-CAR task.

\subsection{Logic-Driven Constraints}
\label{3.2}

In this section, we aim to guide ZS-CAR models with compositional reasoning~\cite{zerroug2022benchmark, gupta2023visual} and fine-to-coarse reasoning~\cite{fei2022making} abilities. The former ability explores interdependent constraints within the architecture of compositions, while the latter adopts the semantic hierarchy to capture inter-class relations between primitives. In spite of previous efforts concerning compositional relationships and semantic primitive hierarchy~\cite{li2024c2c, guo2023texts}, they either concentrate exclusively on composition generation in the test phase while paying inadequate attention to composition structure learning during the training period, or narrowly aim to study the intrinsic semantics of every primitive and remain blind to the rich semantic relationships between different primitives. To address these observed shortcomings, we design two logic-driven constraints: Explicit Compositional Logic for enhancing compositional reasoning ability, and Hierarchical Primitive Logic for fine-to-coarse reasoning ability. Figure~\ref{fig:overview} showcases an overview of LogicCAR.

Our logic-driven constraints are built upon first-order logic~\cite{barwise1977introduction, fischer2019dl2}, a knowledge representation language to express propositional logic. It contains four basic parts: 
\textbf{i)} \textit{constants}, \ie, specific data samples $x_1, x_2, ...$;
\textbf{ii)} \textit{variables} to denote a specific type of constants, \eg, $x$;
\textbf{iii)} \textit{unary predicates} that evaluate the truth values of the semantics of variables (\eg, $\textbf{fall\_likeafeather\_napkin}(x)=true$ indicates that the sample $x$ belongs to the composition category ``\textit{napkin fall like a feather}'');
\textbf{iv)} \textit{connectives} (\eg, $\wedge,\vee,\neg, \Rightarrow$) and \textit{quantifiers} (\ie, $\exists, \forall$) denoting logical relationships.

\begin{figure*}[t]
   \vspace{-10pt}
   \begin{center}
      \includegraphics[width=\linewidth]{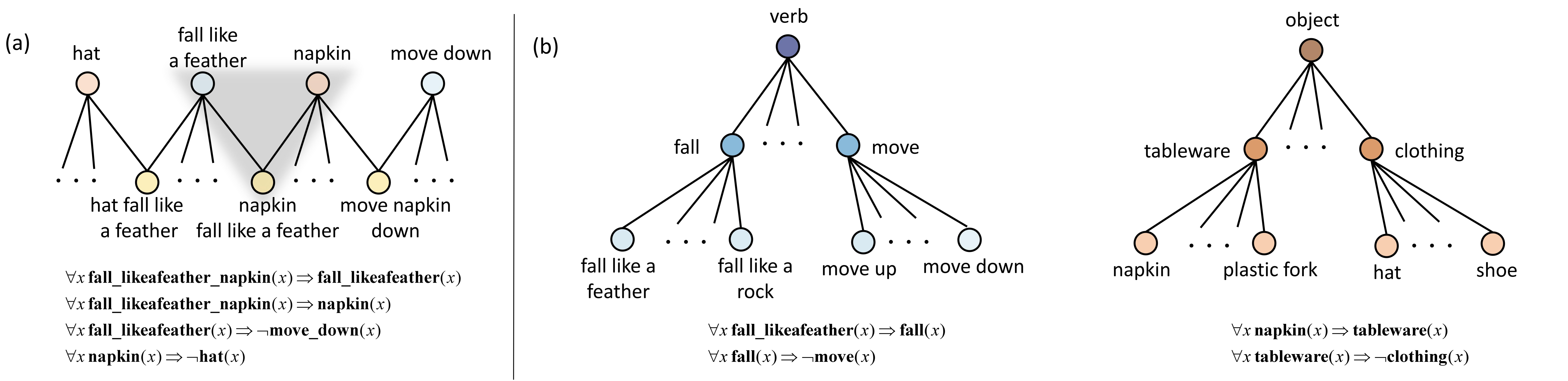}
   \end{center}
   \vspace{-1.0em}
   \captionsetup{font=small}
   \caption{Illustrations of (a) compositional structure constraints; (b) the semantic hierarchy of verb categories and that of object categories.}
   \vspace{-5pt}
   \label{fig:hierarchy}
\end{figure*}

\noindent\textbf{1) {Explicit Compositional Logic.}}
As showcased in Figure~\ref{fig:hierarchy}(a), Explicit Compositional Logic focuses on conditional constraints on the compositional structures. 
In the establishment of our logic rules, we decide to describe compositional structures in the form of logical expressions. Firstly, we argue that a composition inherently embodies a subordinate relationship, \ie, every composition category both belongs to its associated verb and object categories (\textit{composed} relationships), which can be deduced to:
\begin{equation}
\label{composition_c}
\begin{aligned}\small
\forall x(c(x)\Rightarrow v(x)),\\
\forall x(c(x)\Rightarrow o(x)).
  \end{aligned}
\end{equation}
This rule can effectively mitigate our model's tendency to erroneously combine high-frequency primitives. For instance, even if the dataset contains numerous matched samples for both verb category ``fall like a rock'' and object category ``napkin'', LogicCAR does not attribute any special relationship between them, since $\forall x(\textbf{fall\_likearock\_napkin}(x)\Rightarrow {\textbf{fall\_likearock}}(x))$ and $\forall x(\textbf{fall\_likearock\_napkin}(x)\Rightarrow \textbf{napkin}(x))$ are fundamentally nonexistent rules within Explicit Compositional Logic.

Besides, it is also apparent that all verb or object categories exhibit mutual exclusivity (\ie, \textit{exclusive} relationships), leading to:
\begin{equation}
\label{composition_e}
\begin{aligned}\small
\forall x(v(x)\Rightarrow \neg v^1_v(x) \wedge \neg v^2_v(x) \wedge ... 
    \wedge \neg v^{M1}_v(x)), \\
\forall x(o(x)\Rightarrow \neg o^1_o(x) \wedge \neg o^2_o(x) \wedge ... 
    \wedge \neg o^{M2}_o(x)).
  \end{aligned}
\end{equation}
In Eq.~\ref{composition_e}, $v^m_v \in {S_v}$ and $o^m_o \in S_o$ satisfy the following relational expressions: $S_v \cup v = V, S_o \cup o = O$, \ie, $S_v$ and $S_o$ represent all other verb categories for a given verb category $v$ and all other object categories for a given object category $o$, respectively. Take the object category ``\textit{napkin}'' as an example, this rule should be interpreted as ``a napkin is not a hat, a plastic fork ... nor a shoe'', which can be transformed into the expression: $\forall x(\textbf{napkin}(x) \Rightarrow \neg \textbf{hat}(x) \wedge \neg \textbf{plastic\_fork}(x) \wedge ... \wedge \neg \textbf{shoe}(x))$.

\noindent\textbf{2) Hierarchical Primitive Logic.}
As illustrated in Figure~\ref{fig:hierarchy}(b), Hierarchical Primitive Logic explores semantical relations of verb categories and object categories. 
Firstly, to figure out specific dependencies among verb categories, we cluster all verbs based on their semantic features and generate coarse verb categories.
We observe that verb categories all have a key action word and some verb categories share the same action word (\eg, ``\textit{fall}'' for ``\textit{fall like a feather}'' and ``\textit{fall like a rock}''), which enables them to be easily divided into several groups. We classify verb categories with the same action word into one cluster of verbs (\ie, coarse verb category), with the action word as the name of the coarse verb category, \cf, Figure~\ref{fig:hierarchy}(b).
As for object categories, we perform operations similar to those applied to verb categories, \ie, dividing object categories into several coarse object categories according to their semantic similarities. Considering that LLMs possess extensive world knowledge which spans visual and linguistic domains, we employ LLM~\cite{deepseekR1} to do the object classification work. Specifically, we design a customized prompt\footnote{More details about the prompt for LLM are left in the \textbf{Appendix}.} to obtain more reasonable responses from LLM. The object classification results are showcased in Figure~\ref{fig:hierarchy}(b).

In the establishment of logic rules, we highlight the semantic hierarchy of verb categories and object categories. First and foremost, a verb or an object belongs to its coarse category (\textit{composed} relationships), which can be expressed as:
\begin{equation}
\label{hierarchy_c}
\begin{aligned}\small
\forall x(v(x) \Rightarrow \hat{v}(x)), \\
\forall x(o(x) \Rightarrow \hat{c}(x)),
  \end{aligned}
\end{equation}
where $\hat{v}$ and $\hat{c}$ denote the coarse category that contains the given verb $v$ and object $o$, respectively. 
Moreover, there are conflicts among different coarse categories, \ie, one coarse verb category cannot be another coarse verb category at the same time. The \textit{exclusive} relationships can lead to:
\begin{equation}
\label{hierarchy_e}
\begin{aligned}\small
\forall x(\hat{v}(x)\Rightarrow \neg \hat{v}^1_{\hat{v}}(x) \wedge \neg \hat{v}^2_{\hat{v}}(x) \wedge ... 
    \wedge \neg \hat{v}^{N1}_{\hat{v}}(x)), \\
\forall x(\hat{o}(x)\Rightarrow \neg \hat{o}^1_{\hat{o}}(x) \wedge \neg \hat{o}^2_{\hat{o}}(x) \wedge ... 
    \wedge \neg \hat{o}^{N2}_{\hat{o}}(x)), 
  \end{aligned}
\end{equation}
where $\hat{v}^n_{\hat{v}}(x) \in S_{\hat{v}}$ and $\hat{o}^n_{\hat{o}}(x) \in S_{\hat{o}}$ refer to all sibling nodes for a coarse verb category $\hat{v}$ in the semantic hierarchy of verbs and a coarse object category $\hat{o}$ in the semantic hierarchy of objects, respectively.

With specific logic rules determined, we investigate how to transform the aforementioned abstract logical expressions into computable mathematical formulas. Thanks to fuzzy logic~\cite{kosko1993fuzzy, van2022analyzing}, a form of soft probabilistic logic, logic connectives (\eg, $\wedge,\vee,\neg, \Rightarrow$) can be approximately equal to fuzzy operators (\eg, \textit{t-norm}, \textit{t-conorm}, \textit{fuzzy negation}, \textit{fuzzy implication}). In accordance with Goguen fuzzy logic~\cite{hajek2013metamathematics} and Gödel fuzzy logic~\cite{feferman1998kurt}, the precise correspondences are illustrated in the following:
\begin{equation}\small\label{eq:transfer1}
\begin{aligned}\small
   \phi \wedge \varphi \coloneqq \phi \cdot \varphi,~~~ \phi \vee \varphi \coloneqq \max(\phi,\varphi), \\
   ~~~ \neg\phi \coloneqq 1 - \phi, ~~~ \phi\Rightarrow \varphi \coloneqq 1-\phi+\phi\cdot \varphi.
\end{aligned}
\end{equation}
Furthermore, the existential quantifier $\exists$ and the universal quantifier $\forall$ are roughly equivalent to a form of generalized mean~\cite{badreddine2022logic, van2022analyzing}:
\begin{equation}\label{eq:transfer2}
\begin{aligned}\small
\exists x\phi(x)=~& \textstyle(\frac{1}{K}\!\textstyle\sum\nolimits_{k=1}^K \phi(x_k)^q)^\frac{1}{q}, \\
\forall x\phi(x)=~& 1\!-\!\textstyle(\frac{1}{K}\!\textstyle\sum\nolimits_{k=1}^K (1\!-\!\phi(x_k))^q)^\frac{1}{q},
  \end{aligned}
\end{equation}
where $q\in\mathbb{Z}$. As a consequence, our logic rules can be effortlessly transformed into symbolic numerical representations.
For \textit{composed} relationships in Explicit Compositional Logic, Eq.~\ref{composition_c} is mathematically equivalent to:
\begin{equation}\label{eq:C-c}
\begin{aligned}\small
\mathcal{G}_{C\textit{1}}(c)\!=1\!-\!\textstyle\Big(\frac{1}{K}\!\sum_{k=1}^K(\bm{s}_k[c] \!-\! \bm{s}_k[c]\cdot\bm{s}_k[v])^{q}\Big)^{\!\frac{1}{q}} \\
    + 1\!-\!\textstyle\Big(\frac{1}{K}\!\sum_{k=1}^K(\bm{s}_k[c] \!-\! \bm{s}_k[c]\cdot\bm{s}_k[o])^{q}\Big)^{\!\frac{1}{q}},
  \end{aligned}
\end{equation}
{where $\bm{s}_k[c] \in [0,1]$ represents the score (confidence) of $x_k$ for the composition category $c$.}
As for \textit{exclusive} relationships in Explicit Compositional Logic, we convert Eq.~\ref{composition_e} into:
\begin{equation}\label{eq:C-e}
\left\{
\begin{aligned}\small
\mathcal{G}_{C\textit{2}}(v)\!=1\!-\frac{1}{M1}\!\sum_{m=1}^{M1}\!\textstyle\Big(\frac{1}{K}\!\sum_{k=1}^K(\! \bm{s}_k[v]\cdot\bm{s}_k[v^{m}_{v}])^{q}\Big)^{\!\frac{1}{q}}, \\
\mathcal{G}_{C\textit{2}}(o)\!= 1\!-\frac{1}{M2}\!\sum_{m=1}^{M2}\!\textstyle\Big(\frac{1}{K}\!\sum_{k=1}^K(\! \bm{s}_k[o]\cdot\bm{s}_k[o^{m}_{o}])^{q}\Big)^{\!\frac{1}{q}}.
  \end{aligned}
  \right.
\end{equation}
Similar to Eq.~\ref{composition_c}, \textit{composed} relationships in Hierarchical Primitive Logic (\ie, Eq.~\ref{hierarchy_c}) can be derived as:
\begin{equation}\label{eq:H-c}
\left\{
\begin{aligned}\small
\mathcal{G}_{H\textit{1}}(v)\!=1\!-\!\textstyle\Big(\frac{1}{K}\!\sum_{k=1}^K(\bm{s}_k[v] \!-\! \bm{s}_k[v]\cdot\bm{s}_k[\hat{v}])^{q}\Big)^{\!\frac{1}{q}}, \\
\mathcal{G}_{H\textit{1}}(o)\!= 1\!-\!\textstyle\Big(\frac{1}{K}\!\sum_{k=1}^K(\bm{s}_k[o] \!-\! \bm{s}_k[o]\cdot\bm{s}_k[\hat{o}])^{q}\Big)^{\!\frac{1}{q}}.
  \end{aligned}
    \right.
\end{equation}
Lastly, we reformulate Eq.~\ref{hierarchy_e} (\ie, \textit{exclusive} relationships in Hierarchical Primitive Logic) as:
\begin{equation}\label{eq:H-e}
\left\{
\begin{aligned}\small
\mathcal{G}_{H\textit{2}}(\hat{v})\!=1\!-\frac{1}{N1}\!\sum_{n=1}^{N1}\!\textstyle\Big(\frac{1}{K}\!\sum_{k=1}^K(\! \bm{s}_k[\hat{v}]\cdot\bm{s}_k[\hat{v}^{n}_{\hat{v}}])^{q}\Big)^{\!\frac{1}{q}}, \\
\mathcal{G}_{H\textit{2}}(\hat{o})\!=1\!-\frac{1}{N2}\!\sum_{n=1}^{N2}\!\textstyle\Big(\frac{1}{K}\!\sum_{k=1}^K(\! \bm{s}_k[\hat{o}]\cdot\bm{s}_k[\hat{o}^{n}_{\hat{o}}])^{q}\Big)^{\!\frac{1}{q}}.
  \end{aligned}
      \right.
\end{equation}
With the help of the above equations, we can easily set an optimization target in the training for LogicCAR.

\subsection{Training Objectives} 
\label{3.3}
During training, three losses are used to facilitate iterative model optimization:  
a classification loss $\mathcal{L}_{c}$, 
logic losses $\mathcal{L}_{ECL}$ and $\mathcal{L}_{HPL}$ for Explicit Compositional Logic and Hierarchical Primitive Logic. Combining all components, the final loss function takes the form:
\begin{equation}\label{eq:Loss}
\begin{aligned}\small
\mathcal{L} = \mathcal{L}_{c} + \alpha (\mathcal{L}_{ECL}+\beta \mathcal{L}_{HPL}),
  \end{aligned}
\end{equation}
where $\alpha$ and $\beta$ are optimally-chosen coefficients to regulate the impacts of our logic-driven constraints on the training. 

The classification loss $\mathcal{L}_c$ comes from the cross-entropy loss of prediction scores ($\bm{s}$) and the groundtruth (${y}$), calculated as:
\begin{equation}\label{eq:Loss_classification}
\begin{aligned}\small
\mathcal{L}_c = \frac{1}{K}\!\textstyle\sum\nolimits_{k=1}^K \mathcal{L}_{CE}(\bm{s}_k, {y}_k).
  \end{aligned}
\end{equation}

The logic loss $\mathcal{L}_{ECL}$ pushes LogicCAR to further explore the relationships within the structure of seen compositions, effectively reducing the feasibility of impractical combinations. It contains an asymmetric loss $\mathcal{L}_{ea}$ and a rule loss $\mathcal{L}_{er}$:
\begin{equation}\label{eq:Loss-ECL}
\begin{aligned}\small
\mathcal{L}_{ECL} = \mathcal{L}_{ea} + \mathcal{L}_{er}.
  \end{aligned}
\end{equation}
Notably, the asymmetric loss $\mathcal{L}_{ea}$ measures the deviation of our model's predictions on compositions from actual labels.
Through Eq.~\ref{eq:C-c} and Eq.~\ref{eq:C-e}, we arrive at the computation formula for $\mathcal{L}_{er}$:
\begin{equation}\label{eq:Loss-er}
\begin{split}
\mathcal{L}_{er} =\!\textstyle\frac{1}{|\mathcal{N}_{c}|}\!\!\textstyle\sum_{c\in\mathcal{N}_{c}}\!(1-\mathcal{G}_{C\textit{1}}(c)) + \!\textstyle\frac{1}{|\mathcal{N}_{v}|}\!\!\textstyle\sum_{v\in\mathcal{N}_{v}}\!(1-\mathcal{G}_{C\textit{2}}(v)) \\
    +\!\textstyle\frac{1}{|\mathcal{N}_{o}|}\!\!\textstyle\sum_{o\in\mathcal{N}_{o}}\!(1-\mathcal{G}_{C\textit{2}}(o)),
    \end{split}
\end{equation}
where $|\mathcal{N}_{c}|$, $|\mathcal{N}_{v}|$ and $|\mathcal{N}_{o}|$ denote the number of all composition, verb and object categories, respectively.

The logic loss $\mathcal{L}_{HPL}$ optimizes LogicCAR's ability to understand the semantic hierarchy of verb categories and object categories, attaining an improved comprehension of the features of compositional actions. Similar to $\mathcal{L}_{ECL}$, $\mathcal{L}_{HPL}$ also consists of an asymmetric loss $\mathcal{L}_{ha}$ and a rule loss $\mathcal{L}_{hr}$: 
\begin{equation}\label{eq:Loss-HPL}
\begin{aligned}\small
\mathcal{L}_{HPL} = \mathcal{L}_{ha} + \mathcal{L}_{hr}.
  \end{aligned}
\end{equation}
The asymmetric loss calculates the discrepancy between LogicCAR's predictions on verb categories and ground-truth verb labels, as well as the difference between object category predictions and actual object labels.
As for the rule loss $\mathcal{L}_{hr}$, its mathematical formulation is given by:
\begin{equation}\label{eq:Loss-hr}
\begin{split}
\mathcal{L}_{hr} =\!\textstyle\frac{1}{|\mathcal{N}_{v}|}\!\!\textstyle\sum_{v\in\mathcal{N}_{v}}\!(1-\mathcal{G}_{H\textit{1}}(v)) + \!\textstyle\frac{1}{|\mathcal{N}_{o}|}\!\!\textstyle\sum_{o\in\mathcal{N}_{o}}\!(1-\mathcal{G}_{H\textit{1}}(o)) \\
    +\!\textstyle\frac{1}{|\mathcal{N}_{\hat{v}}|}\!\!\textstyle\sum_{\hat{v}\in\mathcal{N}_{\hat{v}}}\!(1-\mathcal{G}_{H\textit{2}}(\hat{v}))
        +\!\textstyle\frac{1}{|\mathcal{N}_{\hat{o}}|}\!\!\textstyle\sum_{\hat{o}\in\mathcal{N}_{\hat{o}}}\!(1-\mathcal{G}_{H\textit{2}}(\hat{o})),
    \end{split}
\end{equation}
where $|\mathcal{{N}}_{\hat{v}}|$ and $|\mathcal{{N}}_{\hat{o}}|$ denote the number of all coarse categories in the semantic hierarchy of verbs and objects, respectively.

\begin{table*}[!t]
  \centering
  \caption{Evaluation results (\%) of CLIP-based models on the test set of Sth-com dataset~\cite{li2024c2c}. \ding{62} indicates that the corresponding experimental results come from existing papers~\cite{li2024c2c}.}
  \vspace{-1.0em}
    \setlength{\tabcolsep}{17pt}
   \renewcommand{\arraystretch}{1.0}
    \begin{tabular}{|r|l||cccccc|}
    \hline\thickhline
    \rowcolor{mygray}
     &  &\multicolumn{6}{c|}{Top-1}     \\
         \rowcolor{mygray}\multirow{-2}[-1]{*}{Method} & \multirow{-2}[-1]{*}{Venue} 
          & \multicolumn{1}{c}{Verb} & \multicolumn{1}{c}{Object} & \multicolumn{1}{c}{Seen} & \multicolumn{1}{c}{Unseen} & \multicolumn{1}{c}{HM} & \multicolumn{1}{c|}{AUC}  \\
    \hline\hline
    CLIP~\cite{radford2021learning}  & ICML’21   &   37.0    &   53.4    &    39.1   &   27.9    &    24.6   &     9.1         \\
    CoOp~\cite{zhou2022conditional} \ding{62} & CVPR’22   &   48.4    &   55.1    &   43.9    &   46.4    &   36.6    &   18.1           \\
    CSP~\cite{nayak2022learning} \ding{62} & ICLR'23   &   49.0    &   54.8    &   43.6    &   47.4    &   36.0    &   18.0           \\    
    DFSP~\cite{lu2023decomposed} \ding{62} & CVPR'23   &   47.4    &   55.4    &   43.1    &   47.3    &   35.8    &   17.9           \\        
    C2C~\cite{li2024c2c} & ECCV'24   &   57.4    &  58.5     &   51.4    &   55.1    &  44.8     & 25.9             \\
    \textbf{LogicCAR} (\textbf{Ours})  &  &   \textbf{57.9}     &   \textbf{59.6}    &  \textbf{51.5}     &  \textbf{57.9}     & \textbf{45.2}     &    \textbf{27.0}            \\ \hline

    \end{tabular}%
  \label{tab:sota}%
    \vspace{-5pt}
\end{table*}%

\section{Experiments}
\subsection{Experimental Setup}
\label{4.1}

\begin{table*}[!t]
  \centering
  \caption{Evaluation results (\%) of logicCAR with different prompting strategies on the test set of Sth-com dataset.}
  \vspace{-1.0em}
    \setlength{\tabcolsep}{7pt}
   \renewcommand{\arraystretch}{1.0}
    \begin{tabular}{|c|c||cccccc|cccc|}
    \hline\thickhline
    \rowcolor{mygray}
     &    &\multicolumn{6}{c|}{Top-1}       & \multicolumn{4}{c|}{Top-5} \\
         \rowcolor{mygray}
         \multirow{-2}[-1]{*}{Method} & 
         \multirow{-2}[-1]{*}{Prompting Strategy} 
          & \multicolumn{1}{c}{Verb} & \multicolumn{1}{c}{Object} & \multicolumn{1}{c}{Seen} & \multicolumn{1}{c}{Unseen} & \multicolumn{1}{c}{HM} & \multicolumn{1}{c|}{AUC}  & \multicolumn{1}{c}{Seen} & \multicolumn{1}{c}{Unseen} & \multicolumn{1}{c}{HM} & \multicolumn{1}{c|}{AUC} \\
    \hline\hline
     \multirow{4}[0]{*}{C2C} & Hard Prompts  & 57.3 & 58.9 & 50.4 & 55.3 & 44.3 & 25.5  & 75.0 & 78.6 & 69.5 & 54.8  \\    
     & SPM~\cite{lu2023decomposed} & 57.3 & 58.2 & 51.2 & 54.5 & 44.4 & 25.5  & 74.5 & 77.2 & 68.9 & 54.9     \\ 
     & CSP~\cite{nayak2022learning} & 57.1 & 58.6 & 50.4 & 54.5 & 44.1 & 25.1   & 74.5 & 78.0 & 69.1 & 55.8      \\    
     & CoOp~\cite{zhou2022conditional}  & 57.4 & 58.5 & 51.4 & 55.1 & 44.8 & 25.9 & 74.4 & 78.3 & 69.7 & 56.1         \\ \hline
     \multirow{4}[0]{*}{LogicCAR (Ours)} & Hard Prompts  &57.7    &    59.6   &   51.1    &   56.4    &    44.8  &    26.2   &  75.4
        &  80.5    &   70.6    &  58.4         \\    
     & SPM~\cite{lu2023decomposed}  &{57.5}     &   {59.5}    &  \textbf{51.7}     &  {56.1}     & {44.8}     &    {26.3}   &  
    \textbf{75.8}    &   {80.1}   &  {70.8}     &  {58.5}          \\ 
     & CSP~\cite{nayak2022learning}  &{57.9}     &   {59.3}    &  {51.2}     &  {55.6}     & {45.2}     &    {26.1}   &  
    {75.4}    &   {80.2}   &  {70.7}     &  {58.2}          \\    
     & CoOp~\cite{zhou2022conditional}   &\textbf{57.9}     &   \textbf{59.6}    &  {51.5}     &  \textbf{57.9}     & \textbf{45.2}     &    \textbf{27.0}   &  
    {75.4}    &   \textbf{81.2}   &  \textbf{71.4}     &  \textbf{59.0}          \\ \hline
    \end{tabular}%
  \label{tab:prompts}%
    \vspace{-5pt}
\end{table*}%

\noindent\textbf{Dataset.}
We performed our experiments on the benchmark dataset for ZS-CAR task: Something-composition (Sth-com) dataset~\cite{li2024c2c}. It is a large-scale dataset which is primarily based on the Something-Something V2 (Sth-V2) dataset~\cite{goyal2017something}. Sth-com dataset comprises 5124 compositions and 79,465 video samples, with 161 verbs and 248 objects. The training set contains 3451 compositions, while the validation/test set includes 733/976 seen compositions and 717/956 unseen compositions.

\noindent\textbf{Evaluation Metrics.} 
We followed the ZS-CAR evaluation protocol~\cite{li2024c2c} and measured all experimental results in six metrics: 1) \textbf{\textit{Verb}}: This metric assesses the accuracy of verbs in compositions. 2) \textbf{\textit{Object}}: This metric shows the accuracy of objects in compositions. 3) \textbf{\textit{Seen}}: This metric evaluates the best accuracy of seen compositions. 4) \textbf{\textit{Unseen}}: This metric measures the best accuracy of unseen compositions. 5) \textbf{\textit{Harmonic Mean} (\textbf{HM})}: The best harmonic mean takes the accuracies of both seen and unseen compositions into consideration, offering a more comprehensive assessment of the performance. 6) \textbf{\textit{Area Under the Curve} (\textbf{AUC})}: This metric calculates the area under the seen-unseen accuracy curve.

\noindent\textbf{Baselines.} We compared LogicCAR with five CLIP-based baseline models: 
1) \textbf{CLIP}, which applies contrastive learning to the cross-modal tasks and utilizes hard prompts (\ie, unlearnable textual inputs) for text encoders.
2) \textbf{CoOp}, which proposes the context optimization to adapt CLIP-like framework for downstream tasks. They design soft prompts (\ie, learnable context vectors) to make the model a data-efficient learner, compared with CLIP.
3) \textbf{CSP}, which fine-tunes the vocabulary for two types of primitives and composes novel soft prompts for unseen composition categories. It separates the prompt optimization of two kinds of primitives, different from CoOp.
4) \textbf{DFSP}, which integrates the approaches of soft prompts and decomposed fusion. Compared with CSP, DFSP decomposes two kinds of primitives among textual features and then fuses them with visual features, respectively. It improves the performance on unseen composition categories and enhances the generalization ability.
5) \textbf{C2C}, which devises a dual-way framework with an enhanced training strategy for ZS-CAR task. It designs \textit{Independent Component Learning} module and \textit{Component to Composition} module to utilize the restrictions within compositions. The enhanced training strategy uses interdependence losses to reduce spurious information and generates synthetic compositions to balance seen and unseen composition learning.

\begin{figure*}
    \centering
    \includegraphics[width=0.95\linewidth]{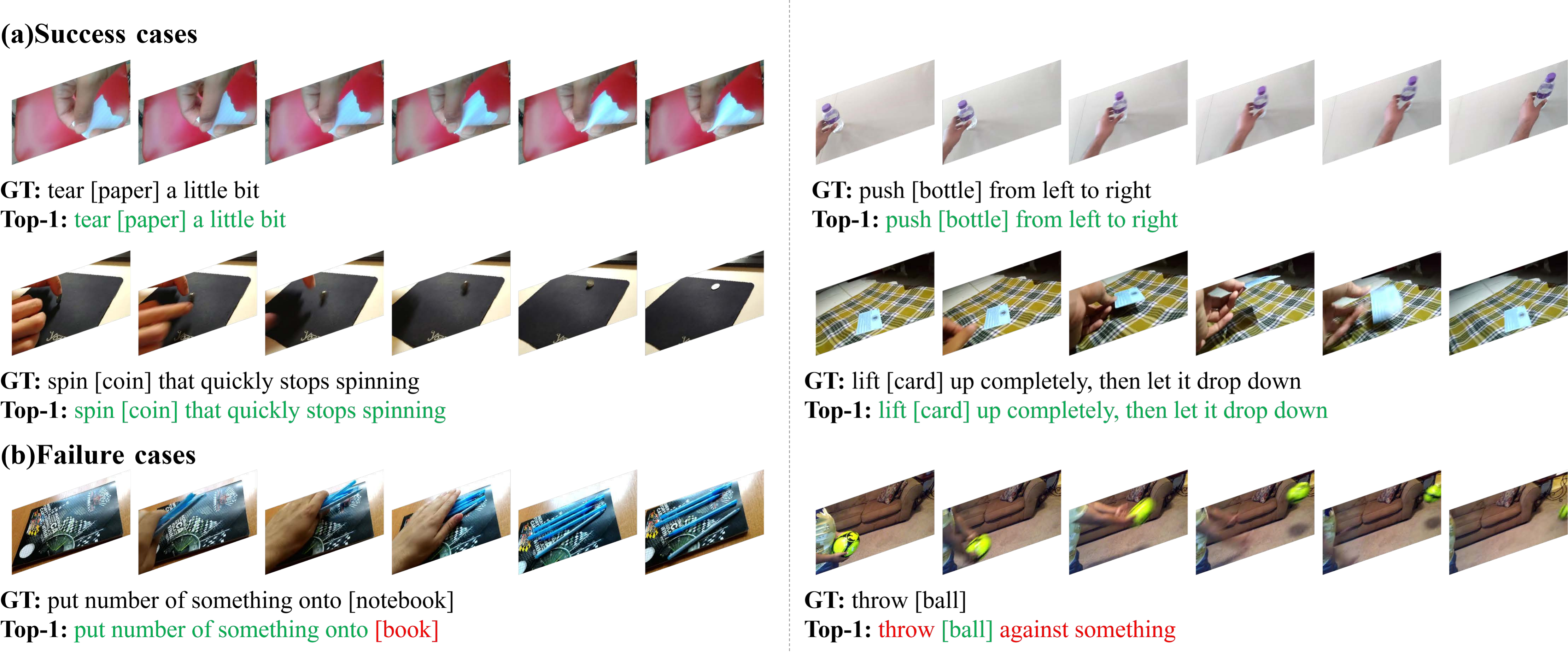}
    \vspace{-1.0em}
    \caption{Top-1 predictions on randomly selected cases from the test set of Sth-com dataset. For Success cases, \ie, (a), the first row shows the results of seen compositions, while the second row displays the performance on unseen compositions. As for Failure cases, \ie, (b), the sample in the left column belongs to a seen composition, while the right one is an unseen composition. Correct and incorrect predictions are highlighted in \textcolor{mygreen}{green} and \textcolor{myred}{red}, respectively.}
    \label{fig:cases}
    \vspace{-5pt}
\end{figure*}
\subsection{Quantitative Comparison Results}
\label{4.2}
We compared the performance of LogicCAR and five CLIP-based baseline models on the test set of Sth-com dataset and report them in Table~\ref{tab:sota}. From the experimental results in Table~\ref{tab:sota}, we observe that:
1) CLIP which provides the fundamental pipeline for ZS-CAR task, demonstrates limited effectiveness owing to the lack of task-oriented optimizations.
2) CoOp exhibits obvious improvements across all metrics compared to CLIP. This is largely due to its enhanced prompting strategy: modeling prompts' context words with learnable vectors.
3) In the \textit{Verb} and \textit{Unseen} metrics, CSP outperforms CoOp by a narrow margin since it separates the optimizations of verb and object primitives.
4) DFSP lags behind CoOp and CSP under the metric of \textit{Seen}, appearing to stem from the confusion caused by its decomposed fusion module.
5) C2C achieves a remarkable advancement over previous methods, which can be attributed to its exploration of the complex interdependence among verb and object primitives.
6) LogicCAR attains superior performance compared with existing approaches, \eg, \textbf{4.2\%} and \textbf{5.1\%} over C2C in terms of \textit{AUC} and \textit{Unseen}, establishing new SOTA standards. This strongly validates the effectiveness of our logic-driven constraints.

\vspace{-0.5em}
\subsection{Ablation Study}
\label{4.3}
In this section, we performed ablation studies to explore the impact of different prompting strategies and the contributions of our two logic-driven constraints. More ablation study results are left in the \textbf{Appendix}.

\textbf{Different Prompting Strategies.} We evaluated the influence of different prompting strategies on logicCAR, and reported the results in Table~\ref{tab:prompts}. We fixed our prompt format as ``\textit{a verb of [verb category]}'' for verb prompts and ``\textit{an object of [object category]}'' for object prompts. It is apparent that LogicCAR obtains consistent improvements over the previous SOTA method, which further corroborates the effectiveness and robustness of our logic-driven constraints. Also, we found out that CoOp is the most suitable prompting strategy for LogicCAR in ZS-CAR task.

\begin{table}[!t]
  \centering
  \caption{Effects of each logic-driven constraints in ZS-CAR. }
  \vspace{-1.0em}
    \setlength{\tabcolsep}{5pt}
    \begin{tabular}{|c|c||c|c|}
    \hline\thickhline
    \rowcolor{mygray}
  \multicolumn{1}{|c|}{Explicit}&{Hierarchical}&{Top-1} & {Top-1} \\
   \rowcolor{mygray}   \multicolumn{1}{|c|}{Compositional Logic} &  {Primitive Logic} &
         {HM} & {AUC} \\
    \hline\hline
          &       & \multicolumn{1}{c|}{44.8} & \multicolumn{1}{c|}{25.9} \\
          \ding{51}  &     & \textbf{45.3}      & 26.3 \\
         & \ding{51}   & {44.7}     & {26.0} \\
    \ding{51}     & \ding{51}    & {45.2}      & \textbf{27.0} \\
    \hline
    \end{tabular}%
  \label{tab:contributions}%
  \vspace{-3.5em}
\end{table}%

\textbf{Contributions of our two Logic-driven Constraints.} We investigated the roles of the proposed logic-driven constraints in LogicCAR. From Table~\ref{tab:contributions}, we observed that: 
1) \textbf{Explicit Compositional Logic}: Combining Explicit Compositional Logic with the pipeline of ZS-CAR task improves both the \textit{HM} metric (from 44.8\% to 45.3\%) and the \textit{AUC} metric (from 25.9\% to 26.3\%). It indicates that the utilization of the compositional structure constraint assists in the overall recognition of compositional actions. 
2) \textbf{Hierarchical Primitive Logic}: Integrating Hierarchical Primitive Logic with the pipeline increases the score of the \textit{AUC} metric (from 25.9\% to 26.0\%), which proves the subtle yet helpful role of leveraging the semantic hierarchical constraint.
3) \textbf{Collaborative Impacts of Explicit Compositional Logic and Hierarchical Primitive Logic}: With both logic-driven constraints, LogicCAR obtains the best performance in the \textit{AUC} metric: 27.0\%. This result outperforms both the one with only Explicit Compositional Logic and the one with only Hierarchical Primitive Logic. It highlights the reciprocal enhancement of the two constraints, which enables the model to achieve the optimal state of zero-shot compositional action recognition.

\subsection{Qualitative Results}
\label{4.4}
We offer Top-1 predictions of LogicCAR on the test set of Sth-com dataset in Figure~\ref{fig:cases}.
Our logic-driven constraints make great contributions to understanding the restrictions within compositions and correlations among primitives, which significantly enhance the generalization ability of LogicCAR. As shown in Figure~\ref{fig:cases}, LogicCAR performs well in both seen and unseen composition recognition. It should be noted that the incorrect predictions generated by LogicCAR are not entirely unreasonable. For example, LogicCAR successfully predicts the object category ``\textit{ball}'', but wrongly classifies the verb category ``\textit{throw}'' into the verb category ``\textit{throw ... against something}''. These qualitative results indicate the contributions of our logic-driven constraints in compositional action comprehension of videos.

\section{Conclusion}
In this paper, we discussed the challenges in applying existing compositional learning methods to ZS-CAR task. Inspired by the logical-thinking ability of humans, we proposed our logic reasoning framework LogicCAR with two logic-driven constraints: Explicit Compositional Logic and Hierarchical Primitive Logic, which are based on symbolic reasoning. Empowered by the two logic-driven constraints, LogicCAR conducts an in-depth investigation into the compositional structure constraint and the semantic hierarchy constraint. Therefore, it enhances its compositional reasoning ability and fine-to-coarse reasoning ability. Experimental results on the ZS-CAR benchmark demonstrate LogicCAR outperforms current baseline methods. LogicCAR opens a new avenue for ZS-CAR from the perspective of equipping models with symbolic reasoning, and we hope it will stimulate our community to investigate new methodologies in this field.

\section{Acknowledgments}

This work was supported by the National Natural Science Foundation of China (62441617), Zhejiang Provincial Natural Science Foundation of China (No. LD25F020001), and Fundamental Research Funds for the Central Universities (226-2025-00057). 
This work was also supported by the Hong Kong SAR RGC Early Career Scheme (26208924), the National Natural Science Foundation of China Young Scholar Fund (62402408), and the HKUST Sports Science and Technology Research Grant (SSTRG24EG04).
This research was partially conducted by ACCESS – AI Chip Center for Emerging Smart Systems, supported by the InnoHK initiative of the Innovation and Technology Commission of the Hong Kong Special Administrative Region Government.

\bibliographystyle{ACM-Reference-Format}
\bibliography{acmart}

\renewcommand{\thetable}{A\arabic{table}}
\renewcommand{\thefigure}{A\arabic{figure}}

\clearpage
\setcounter{page}{1}

\twocolumn[{
  \centering
  \huge \textbf{Zero-shot Compositional Action         Recognition with Neural Logic Constraints\\}
  \huge {Supplementary Materials}
  \vspace{1em}  
}]
\setcounter{section}{0}
\setcounter{table}{0}
\setcounter{figure}{0}

\section*{Appendix}
Our supplementary materials are organized as follows:
\begin{itemize}
\item More Related Works are shown in Sec.~\ref{sec:related}.
\item The implementation details are provided in Sec.~\ref{sec:imple}.
\item The object classification generation by LLM are illustrated in Sec.~\ref{sec:ocg}.
\item Explanations for some results \& questions are offered in Sec.~\ref{sec:exp}.
\item Additional ablation study results are presented in Sec.~\ref{sec:aba}.
\item Potential societal impacts are discussed in Sec.~\ref{sec:influence}.
\item Limitations are considered in Sec.~\ref{sec:lim}.
\end{itemize}

\section{More Related Works}
\label{sec:related}
\noindent\textbf{Recent advancements in Compositional Action Recognition (CAR).} Meanwhile, the success of knowledge-based frameworks~\cite{shao2021improving, liu2024knowledge, chen2025vision, shao2024knowledge} in other fields like action recognition inspires researchers to integrate such methods with CAR models. KCMM~\cite{liu2025knowledge} recomposes labels and features to generate unseen compositions, then leverages motion priors from ConceptNet to assess their plausibility. Huang \etal ~\cite{huang2025revisiting} propose a trio-tuning-testing method for few-shot CAR, featuring an inner-to-outer approach for learning verbs \& objects and a Trio-Knowledge Calibration strategy that fuses visual-semantic prototypes to boost generalization.

\noindent\textbf{Zero-shot Compositional Action Recognition (ZS-CAR)}. Unlike the broader CAR task~\cite{materzynska2020something, ji2020action}, which only predicts verb categories during testing, ZS-CAR requires models to recognize both verbs and objects. It is a compositional generalization problem of complex video understanding. Based on Something-Something v2 dataset~\cite{mahdisoltani2018effectiveness}, Li \etal ~\cite{li2024c2c} propose Something-composition (Sth-com) benchmark, containing 5,124 compositions and 79,465 video samples. They also introduce C2C, a framework with two key designs: 1) Component modeling modules to capture composition restrictions; 2) Interdependence losses and synthetic data generation to balance seen and unseen composition learning. However, the framework fails to leverage compositional structure constraint and semantic hierarchy constraint, thus suffering from limited performance.

\section{Implementation Details}
\label{sec:imple}
We implemented LogicCAR using the framework of PyTorch~\cite{pytorch} and trained it on NVIDIA GeForce RTX 3090. We chose the ViT-B/32 transformer as the backbone of CLIP. As for LLM to generate the semantic hierarchy for object categories, we used DeepSeek-R1-Distill-Llama-70B. In the training objective, we assigned our logic coefficients $\alpha$ and $\beta$ to 0.04 and 0.06, respectively. Besides, we set the temperature parameter in our asymmetric loss to 15.0, based on empirical evidence. Furthermore, considering the training instability in early stages, we employ a step-based approach to control the phase of integrating our rule loss into logic losses. Empirically, we set the step as 2 epochs. As for the specific optimization strategy, we followed the ~\cite{li2024c2c} and employed AdamW optimizer for 50 epochs on Sth-com dataset, with the learning rate for parameters in text encoders equal to 1e-4 and the learning rate for parameters in visual encoders equal to 5e-4. Moreover, we chose a batch size of 64 to train our model. To ensure the reproducibility of our method, the code will be made publicly available soon.

\section{Object Classification Generation}
\label{sec:ocg}
We designed a prompt to ask LLM to generate an object classification result for all object categories we provided. The prompt's structure is as follows:
 \lstset{
  basicstyle=\ttfamily,
  columns=fullflexible, 
}
\begin{lstlisting}[numbers=none,breaklines=true]
Q: Categorize the following object into a broad category: chair.
A: chair belongs to furniture.
Complete the following dialog with the format of the example. Do not print any extra words!
Q: Categorize the following object into a broad category: [OBJECT CATEGORY].
A: [OBJECT CATEGORY] belongs to 
\end{lstlisting}

To enhance the rationality and reliability of LLM responses, we utilized the technique of in-context learning~\cite{dong2022survey, menon2022visual, shao2025micas} in our prompt. Before raising the target question, we presented a question-answer pair that exemplifies the expected response structure and content depth. Considering the stochastic nature of LLM responses, we adopted a multi-query approach and obtained 19 independent responses from LLM. 

\begin{figure*}[t]
   \vspace{-10pt}
   \begin{center}
      \includegraphics[width=\linewidth]{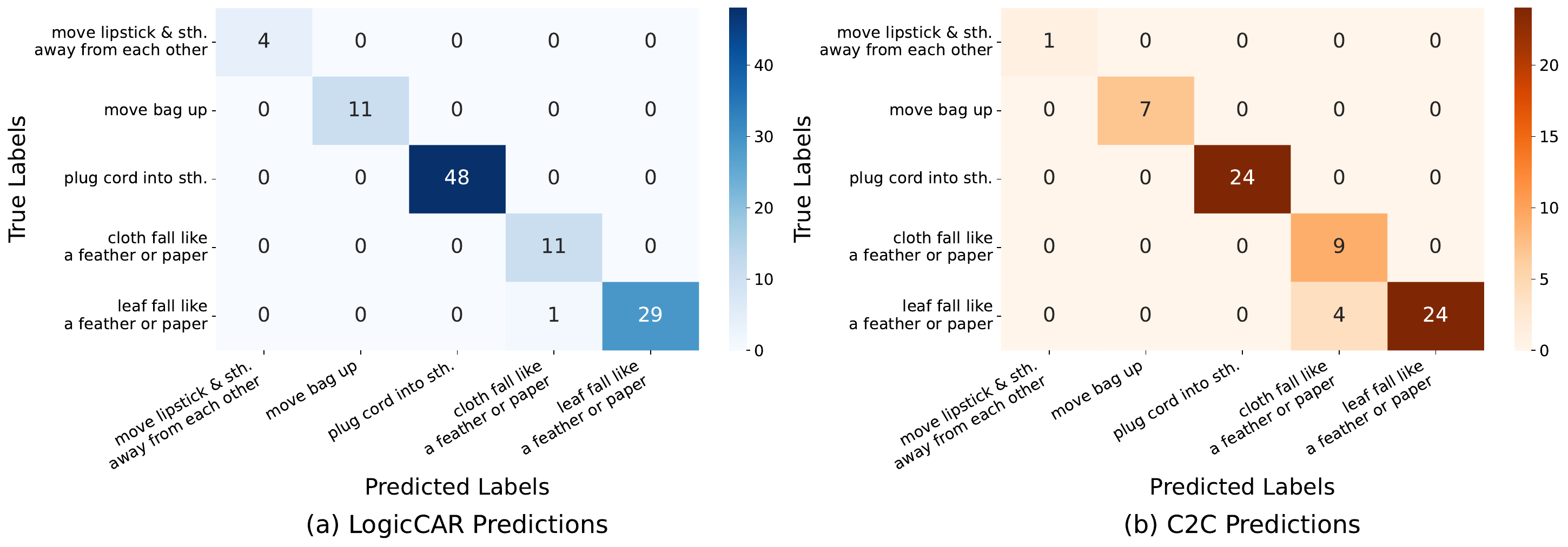}
   \end{center}
   \vspace{-1.0em}
   \captionsetup{font=small}
   \caption{Illustrations of confusion matrices (subset of categories shown) for (a) LogicCAR predictions and (b) C2C predictions.}
   \label{fig:confusion}
\end{figure*}

\section{Explanations for some Results \& Questions}
\label{sec:exp}
\textbf{Explanation for Table 3.} Explicit Compositional Logic enhances LogicCAR's compositional reasoning ability to improve unseen category recognition, while Hierarchical Primitive Logic strengthens its fine-to-coarse reasoning capacity to recognize seen compositions. Due to semantic overlap among both seen labels (\eg, ``put bottle upright on table'') and unseen labels (\eg, ``put bottle on a surface''), it's hard to improve the performance of both simultaneously. HM is a trade-off metric, but may favor unseen ones. Thus, Explicit Compositional Logic gets relatively higher HM, as shown in Table~\ref{tab:effects}. Jointly using two logic constraints, LogicCAR achieves an optimal trade-off among all metrics.

\begin{table}[!t]
  \centering
  \caption{Effects of each logic constraint in ZS-CAR. }
  \vspace{-0.5em}
    \setlength{\tabcolsep}{2.5pt}
   \renewcommand{\arraystretch}{1.0}
    \begin{tabular}{|c|c||cccc|}
    \hline\thickhline
    \rowcolor{mygray}
        \multicolumn{1}{|c|}{Explicit} & \multicolumn{1}{c||}{Hierarchical} & \multicolumn{4}{c|}{Top-1}    \\
         \rowcolor{mygray}

         \multicolumn{1}{|c|}{Compositional Logic} 
         & \multicolumn{1}{c||}{Primitive Logic} &
           \multicolumn{1}{c}{Seen} & \multicolumn{1}{c}{Unseen} & \multicolumn{1}{c}{HM} & \multicolumn{1}{c|}{AUC}  \\
    \hline\hline
      &     & 51.4 & 55.1 & 44.8 & 25.9 \\
      \ding{51}& & 51.1 & 56.4 & \textbf{45.3} & 26.3\\
      & \ding{51} & \textbf{51.9} & 55.0 & 44.7 & 26.0\\
     \ding{51}& \ding{51}   &  {51.5}     &  \textbf{57.9}     & {45.2}     &    \textbf{27.0}        \\ \hline
    \end{tabular}%
  \label{tab:effects}%
\end{table}%

\textbf{Countermeasures for Unreasonable Combinations.}
We design our logic constraints to enhance ZS-CAR models' reasoning abilities: (1) Explicit Compositional Logic prevents structural errors. It defines strict rules based only on seen compositions, which directly mitigates spurious correlations (predicting “fold key” just because they are frequent primitives). The model is penalized via a logic loss for forming such ``illogical combinations''.
(2) Hierarchical Primitive Logic offers semantic common sense. It groups primitives into hierarchical, mutually exclusive coarse categories (\eg, ``chair→furniture'' vs. ``apple→fruit''). Thus combinations are not only structurally valid but also semantically coherent. Empirically, LogicCAR achieves SOTA results over all metrics (\cf, Table 1), proving its enhanced reasoning abilities.

\textbf{Further Enhancement on Model Autonomy.}
While the formulation of logic rules is partly human-guided, which also commonly exists in the current work~\cite{li2023logicseg, liang2023logic}, we reduce manual intervention by: (1) {Employing an LLM} to automatically classify {objects} into coarse categories by utilizing its extensive world knowledge. (2) Using an {inherent heuristic} that clusters {verbs} sharing the same action word (\eg, ``fall'' in ``fall like a feather'' and ``fall like a rock''). Empirically, LogicCAR still achieves SOTA performance, proving its efficacy. Future work will {explore automated constraint learning} and {LLM-generated robust constraints} to further boost autonomy.

\section{Extra Ablation Study}
\label{sec:aba}
\begin{figure}
    \centering
    \includegraphics[width=1\linewidth]{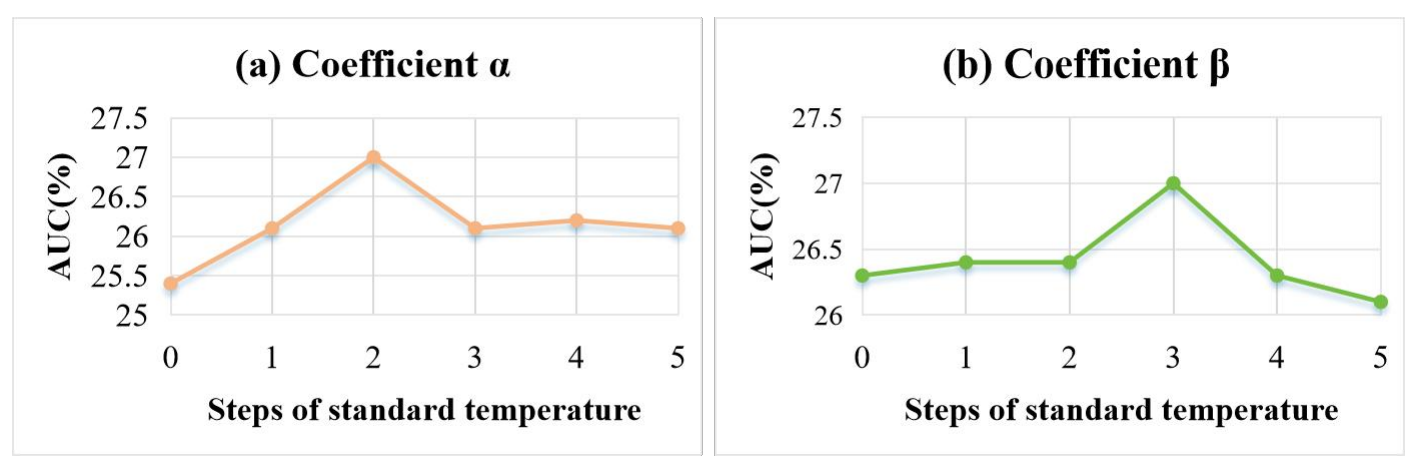}
    \vspace{-2.7em}
    \caption{Ablation study on the coefficients of our logic-driven constraints.}
    \label{fig:coe}
    \vspace{-0.5em}
\end{figure}

\textbf{Coefficients of two Constraints.} We assessed LogicCAR's performance under various logic coefficients (\ie, $\alpha$ and $\beta$ in Eq. ~\ref{eq:Loss}). We empirically set the temperature as 0.02, and compared the impacts of different values of Coefficient $\alpha$ and Coefficient $\beta$. As seen in Figure~\ref{fig:coe} (a), the relationship between the performance of LogicCAR and $\alpha$ follows an inverted $U$-shaped pattern, with the optimal value of \textit{AUC} achieved at $\alpha$ equal to 0.04. Figure~\ref{fig:coe} (b) reveals that the influence of $\beta$ on LogicCAR follows a similar trend with $\alpha$, showing a maximum \textit{AUC} value when $\beta$ = 0.06. However, it should be noted that an excessively large $\beta$ value can lead to a performance drop, since it may \textbf{overemphasize} the role of category clusters and blur critical decision boundaries. 

\textbf{Robustness of LLM responses.}
The evaluation results are depicted in Table~\ref{tab:llm}. Notably, by integrating with different LLM responses, LogicCAR demonstrates a relatively stable performance, \eg, both the AUC metric (26.7\% $\pm$ 0.3\%) and the HM metric (45.0\% $\pm$ 0.2\%) have a narrow range. This underlines the robustness of our LLM responses and the effectiveness of our prompt designed for LLM.

\begin{table}[!t]
  \centering
  \caption{Evaluation results (\%) of LogicCAR with different object classification results on the test set of Sth-com dataset. }
  \vspace{-0.5em}
    \setlength{\tabcolsep}{4.5pt}
   \renewcommand{\arraystretch}{1.0}
    \begin{tabular}{|c||cccccc|}
    \hline\thickhline
    \rowcolor{mygray}
        \multicolumn{1}{|c||}{Number of} & \multicolumn{6}{c|}{Top-1}    \\
         \rowcolor{mygray}
         \multicolumn{1}{|c||}{LLM responses}
          & \multicolumn{1}{c}{Verb} & \multicolumn{1}{c}{Object} & \multicolumn{1}{c}{Seen} & \multicolumn{1}{c}{Unseen} & \multicolumn{1}{c}{HM} & \multicolumn{1}{c|}{AUC}  \\
    \hline\hline

      1    & 57.3 & 59.6 & 51.0 & 57.2 & 45.1 & 26.5 \\
      5 & 57.2 & 59.5 & 51.3 & 56.7 & 44.9 & 26.4\\
      10 & 57.5 & \textbf{59.7} & 51.3 & 56.9 & 44.9 & 26.4\\
     19 (All)  &\textbf{57.9}     &   {59.6}    &  {51.5}     &  \textbf{57.9}     & \textbf{45.2}     &    \textbf{27.0}        \\ \hline
    \end{tabular}%
  \label{tab:llm}%
\end{table}%

\textbf{Confusion Matrices for Final Predictions.}
To explore the effect of logic constraints, we displayed the confusion matrices in Figure~\ref{fig:confusion}. With better separation of similar compositions (\eg, ``cloth fall like a feather or paper'' \& ``leaf fall like a feather or paper''), LogicCAR improves precision-recall performance over C2C, proving its effectiveness.

\textbf{Evaluation Results on Vision Model Backbones.} 
We reported the comparison results between LogicCAR and C2C based on different vision model backbones in Table~\ref{tab:backbones}. LogicCAR demonstrates significant improvements over C2C, \eg, 36.2\% → 37.3\% HM gains under VideoSwin-T backbone.

\begin{table}[!t]
  \centering
  \caption{Evaluation results (\%) of different vision model backbones on Sth-com test set. }
  \vspace{-0.5em}
    \setlength{\tabcolsep}{6pt}
   \renewcommand{\arraystretch}{1.0}
      \begin{tabularx}{\linewidth}{|>{\centering\arraybackslash}X|>{\centering\arraybackslash}X|>{\centering\arraybackslash}X|} 
    \hline\thickhline
    \rowcolor{mygray}
        {Backbones} & {Methods} & {Top-1 HM}\\
    \hline\hline 
      TSM-18&   C2C  & 32.5\\
      TSM-18&   LogicCAR  & \textbf{32.8}\\
      VideoSwin-T & C2C & 36.2\\
      VideoSwin-T & LogicCAR & \textbf{37.3}       \\ \hline
    \end{tabularx}%
  \label{tab:backbones}%
\end{table}%

\textbf{Performance Under Real-world Interference Circumstance.} 
To evaluate LogicCAR's performance under real-world interference, we conducted experiments on two datasets:
(1) Noisy test set of Sth-com: We introduced different types of noises into Sth-com test set. As shown in Table~\ref{tab:inter1}, HM and AUC show small deviations (within \textbf{0.4\%}) across noise conditions.
(2) Real-world Interference dataset: We constructed it by selecting some noisy videos from Sth-com and collecting additional real-world interference samples. Table~\ref{tab:inter2} indicates that LogicCAR outperforms C2C under the challenging conditions, exhibiting its resistance to interference.

\textbf{Model Complexity \& Computational Efficiency.} 
We compared the per-epoch computational costs between LogicCAR and C2C in Table~\ref{tab:time}. Experimental results indicate that our approach maintains comparable computational efficiency to baseline C2C while achieving higher performance. This stems from the fact that our logic constraints serve as loss terms in the whole LogicCAR framework, which won't introduce extra computational overhead.

\begin{table}[!t]
  \centering
  \caption{Evaluation results (\%) of LogicCAR on Noisy test set of Sth-com. }
  \vspace{-0.5em}
    \setlength{\tabcolsep}{4pt}
   \renewcommand{\arraystretch}{1.0}
      \begin{tabularx}{\linewidth}{|>{\centering\arraybackslash}X|>{\centering\arraybackslash}X|>{\centering\arraybackslash}X|} 
    \hline\thickhline
    \rowcolor{mygray}
        {Noise Type} & {Top-1 HM} & {Top-1 AUC}\\
    \hline\hline 
      No Noise&  \textbf{45.2}  & \textbf{27.0}\\
      Gauss Noise&   45.1  & 26.8\\
      Salt \& Pepper Noise & 45.0 & 26.8\\
      Blur & \textbf{45.2} & 26.8       \\ 
      Occlusion & 45.1 & 26.6 \\\hline
    \end{tabularx}%
  \label{tab:inter1}%
\end{table}%

\begin{table}[!t]
  \centering
  \caption{Evaluation results (\%) on Real-world Interference dataset. }
  \vspace{-0.5em}
    \setlength{\tabcolsep}{6pt}
   \renewcommand{\arraystretch}{1.0}
      \begin{tabularx}{\linewidth}{|>{\centering\arraybackslash}X|>{\centering\arraybackslash}X|>{\centering\arraybackslash}X|} 
    \hline\thickhline
    \rowcolor{mygray}
        {Methods} & {Top-1 HM} & {Top-1 AUC}\\
    \hline\hline 
      C2C&  54.5  & 38.0\\
      LogicCAR&  \textbf{55.1}  & \textbf{41.6}\\ \hline
    \end{tabularx}%
  \label{tab:inter2}%
\end{table}%

\begin{table}[!t]
  \centering
  \caption{Per-epoch training \& inference time (s).}
  \vspace{-0.5em}
    \setlength{\tabcolsep}{2.5pt}
   \renewcommand{\arraystretch}{1.0}
      \begin{tabularx}{\linewidth}{|>{\centering\arraybackslash}X|>{\centering\arraybackslash}X|>{\centering\arraybackslash}X|} 
    \hline\thickhline
    \rowcolor{mygray}
        {Methods} & {Training} & {Inference}\\
    \hline\hline 
      C2C&   1406.9  & 249.7\\
      LogicCAR  & 1446.7 & 243.4   \\ \hline
    \end{tabularx}%
  \label{tab:time}%
\end{table}%

\section{Potential societal impacts}
\label{sec:influence}
\textbf{Potential negative societal influence.} In spite of the satisfactory performance of our logic-driven constraints, they may produce unplanned societal effects. Since our work is built upon some foundation models (\eg, pre-trained vision-language models and large language models), there may exist societal biases in our model if the foundation models are trained with imbalanced datasets.

\textbf{Broader impacts.} This work enables a fine-grained understanding of complex human actions, which enhances real-world applications such as intelligent surveillance (\eg, detecting aggressive behaviors), healthcare (\eg, remote rehabilitation monitoring) and so on. By releasing our code, we hope it will encourage further research in bridging the gap between advanced AI and socially meaningful solutions.

\section{Limitations}
\label{sec:lim}
While our work exhibits superior performance in the ZS-CAR task, there still exist two limitations of our logic-driven constraints: 
\textbf{1) Reliance on Third-Party Language Model Services.} Our implementation calls the LLM inference API for object classification, which poses a potential challenge in large-scale deployment. When faced with a dataset that contains a large number of object categories, the costs of API requests may experience significant inflation.
\textbf{2) Boundaries of Zero-Shot Scenarios.} Our work focuses on the recognition of unseen compositions with seen verb categories and seen object categories. It is incapable of identifying compositions with totally novel verbs and objects. This limitation suggests
a viable direction for future improvement.

\end{document}